# Learning-based Video Motion Magnification

Tae-Hyun Oh[1]*, Ronnachai Jaroensri[1]*, Changil Kim[1], Mohamed Elgharib[2],
Frédo Durand[1], William T. Freeman[1,3], and Wojciech Matusik[1]

[1] MIT CSAIL, Cambridge, MA, USA
[2] HBKU QCRI, Doha, Qatar
[3] Google Research
{taehyun, tiam}@csail.mit.edu

**Abstract.** Video motion magnification techniques allow us to see small motions previously invisible to the naked eyes, such as those of vibrating airplane wings, or swaying buildings under the influence of the wind. Because the motion is small, the magnification results are prone to noise or excessive blurring. The state of the art relies on hand-designed filters to extract representations that may not be optimal. In this paper, we seek to learn the filters directly from examples using deep convolutional neural networks. To make training tractable, we carefully design a synthetic dataset that captures small motion well, and use two-frame input for training. We show that the learned filters achieve high-quality results on real videos, with less ringing artifacts and better noise characteristics than previous methods. While our model is not trained with temporal filters, we found that the temporal filters can be used with our extracted representations up to a moderate magnification, enabling a frequency-based motion selection. Finally, we analyze the learned filters and show that they behave similarly to the derivative filters used in previous works. Our code, trained model, and datasets will be available online.

## 1 Introduction

The ability to discern small motions enables important applications such as understanding a building's structural health [3] and measuring a person's vital sign [1]. Video motion magnification techniques allow us to perceive such motions. This is a difficult task, because the motions are so small that they can be indistinguishable from noise. As a result, current video magnification techniques suffer from noisy outputs and excessive blurring, especially when the magnification factor is large [26, 30, 27, 31].

Current video magnification techniques typically decompose video frames into representations that allow them to magnify motion [26, 30, 27, 31]. Their decomposition typically relies on hand-designed filters, such as the complex steerable filters [6], which may not be optimal. In this paper, we seek to learn the

---
*These authors contributed equally.*

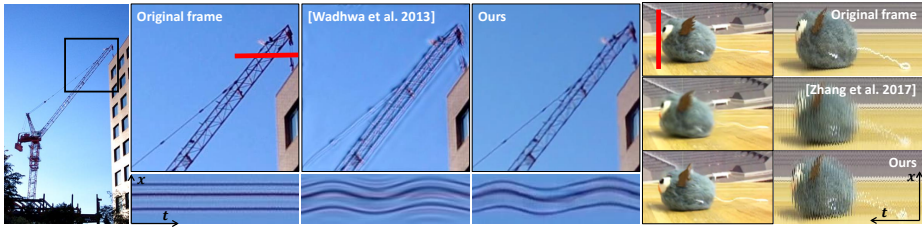

**Fig. 1.** While our model learns spatial decomposition filters from synthetically generated inputs, it performs well on real videos with results showing less ringing artifacts and noise. (Left) the *crane* sequence magnified 75× with the same temporal filter as Wadhwa *et al.* [26]. (Right) `Dynamic` mode magnifies difference (velocity) between consecutive frames, allowing us to deal with large motion as did Zhang *et al.* [31]. The red lines indicate the sampled regions for drawing x-t and y-t slice views.

decomposition filter directly from examples using deep convolutional neural networks (CNN). Because real motion-magnified video pairs are difficult to obtain, we designed a synthetic dataset that realistically simulates small motion. We carefully interpolate pixel values, and we explicitly model quantization, which could round away sub-level values that result from subpixel motions. These careful considerations allow us to train a network that generalizes well in real videos.

Motivated by Wadhwa *et al.* [26], we design a network consisting of three main parts: the spatial decomposition filters, the representation manipulator, and the reconstruction filters. To make training tractable, we simplify our training using two-frame input, and the magnified difference as the target instead of fully specifying temporal aspects of motion. Despite training on the simplified two-frames setting and synthetic data, our network achieves better noise performance and has fewer edge artifacts (See Fig. 1). Our result also suggests that the learned representations support linear operations enough to be used with linear temporal filters up to a moderate magnification factor. This enables us to select motion based on frequency bands of interest.

Finally, we visualize the learned filters and the activations to have a better understanding of what the network has learned. While the filter weights themselves show no apparent pattern, a linear approximation of our learned (non-linear) filters resembles derivative filters, which are the basis for decomposition filters in the prior art [30, 26].

The main contributions of this paper are as follows:

- We present the first learning-based approach for the video motion magnification, which achieves high-quality magnification with fewer ringing artifacts, and has better noise characteristics.
- We present a synthetic data generation method that captures small motions, allowing the learned filters to generalize well in real videos.
- We analyze our model, and show that our learned filters exhibit similarity to the previously hand-engineered filters.

We will release the codes, the trained model, and the dataset online.

| Method | Liu *et al.* [14] | Wu *et al.* [30] | Wadhwa *et al.* [26] | Wadhwa *et al.* [27] | Zhang *et al.* [31] | Ours |
|---|---|---|---|---|---|---|
| **Spatial decomposition** | Tracking, optical flow | Laplacian pyramid | Steerable filters | Riesz pyramid | Steerable filters | Deep convolution layers |
| **Motion isolation** | - | Temporal bandpass filter | Temporal bandpass filter | Temporal bandpass filter | Temporal bandpass filter (2nd-order derivative) | Subtraction or temporal bandpass filter |
| **Representation denoising** | Expectation-Maximization | - | Amplitude weighted Gaussian filtering | Amplitude weighted Gaussian filtering | Amplitude weighted Gaussian filtering | Trainable convolution |

**Table 1. Comparisons of the prior arts.**

## 2  Related Work

**Video motion magnification.** Motion magnification techniques can be divided into two categories: Lagrangian and Eulerian approaches. The Lagrangian approach explicitly extracts the motion field (optical flow) and uses it to move the pixels directly [14]. The Eulerian approaches [30, 26, 27], on the other hand, decompose video frames into representations that facilitate manipulation of motions, without requiring explicit tracking. These techniques usually consist of three stages: decomposing frames into an alternative representation, manipulating the representation, and reconstructing the manipulated representation to magnified frames. Wu *et al.* [30] use a spatial decomposition motivated by the first-order Taylor expansion, while Wadhwa *et al.* [26, 27] use the complex steerable pyramid [6] to extract a phase-based representation. Current Eulerian techniques are good at revealing subtle motions, but they are hand-designed [30, 26, 27], and do not take into account many issues such as occlusion. Because of this, they are prone to noise and often suffer from excessive blurring. Our technique belongs to the Eulerian approach, but our decomposition is directly learned from examples, so it has fewer edge artifacts and better noise characteristics.

One key component of the previous motion magnification techniques is the multi-frame temporal filtering over the representations, which helps to isolate motions of interest and to prevent noise from being magnified. Wu *et al.* [30] and Wadhwa *et al.* [26, 27] utilize standard frequency bandpass filters. Their methods achieve high-quality results, but suffer from degraded quality when large motions or drifts occur in the input video. Elgharib *et al.* [4] and Zhang *et al.* [31] address this limitation. Elgharib *et al.* [4] model large motions using affine transformation, while Zhang *et al.* [31] use a different temporal processing equivalent to a second-order derivative (*i.e.*, acceleration). On the other hand, our method achieves comparable quality even without using temporal filtering. The comparisons of our method to the prior arts are summarized in Table 1.

**Deep representation for video synthesis.** Frame interpolation can be viewed as a complementary problem to the motion magnification problem, where the magnification factor is less than 1. Recent techniques demonstrate high-quality results by explicitly shifting pixels using either optical flow [10, 28, 15] or pixel-shifting convolution kernels [18, 19]. However, these techniques usually require re-training when changing the manipulation factor. Our representation can be directly configured for different magnification factors without re-training. For frame extrapolation, there is a line of recent work [17, 24, 25] that directly synthesizes RGB pixel values to predict dynamic video frames in the future, but their results are often blurry. Our work focusing on magnifying motion within a video, without concerns about what happens in the future.

## 3 Learning-based Motion Magnification

In this section, we introduce the motion magnification problem and our learning setup. Then, we explain how we simplify the learning to make it tractable. Finally, we describe the network architecture and give the full detail of our dataset generation.

### 3.1 Problem statement

We follow Wu *et al.*'s and Wadhwa *et al.*'s definition of motion magnification [30, 26]. Namely, given an image $I(\mathbf{x}, t) = f(\mathbf{x} + \delta(\mathbf{x}, t))$, where $\delta(\mathbf{x}, t)$ represents the motion field as a function of position $\mathbf{x}$ and time $t$, the goal of motion magnification is to magnify the motion such that the magnified image $\tilde{I}$ becomes

$$\tilde{I}(\mathbf{x}, t) = f(\mathbf{x} + (1 + \alpha)\delta(\mathbf{x}, t)), \tag{1}$$

where $\alpha$ is the magnification factor. In practice, we only want to magnify certain signal $\tilde{\delta}(\mathbf{x}, t) = \mathcal{T}(\delta(\mathbf{x}, t))$ for a selector $\mathcal{T}(\cdot)$ that selects motion of interest, which is typically a temporal bandpass filter [26, 30].

While previous techniques rely on hand-crafted filters [26, 30], our goal is to learn a set of filters that extracts and manipulates representations of the motion signal $\delta(\mathbf{x}, t)$ to generate output magnified frames. To simplify our training, we consider a simple two-frames input case. Specifically, we generate two input frames, $\mathbf{X}_a$ and $\mathbf{X}_b$ with a small motion displacement, and an output motion-magnified frame $\mathbf{Y}$ of $\mathbf{X}_b$ with respect to $\mathbf{X}_a$. This reduces parameters characterizing each training pair to just the magnification factor. While this simplified setting loses the temporal aspect of motion, we will show that the network learns a linear enough representation *w.r.t.* the displacement to be compatible with linear temporal filters up to a moderate magnification factor.

### 3.2 Deep Convolutional Neural Network Architecture

Similar to Wadhwa *et al.* [26], our goal is to design a network that extracts a representation, which we can use to manipulate motion simply by multiplication and to reconstruct a magnified frame. Therefore, our network consists of three parts: the encoder $G_e(\cdot)$, the manipulator $G_m(\cdot)$, and the decoder $G_d(\cdot)$, as illustrated in Fig. 2. The encoder acts as a spatial decomposition filter that extracts a shape representation [9] from a single frame, which we can use to manipulate motion (analogous to the phase of the steerable pyramid and Riesz pyramid [26, 27]). The manipulator takes this representation and manipulates it to magnify the motion (by multiplying the difference). Finally, the decoder reconstructs the modified representation into the resulting motion-magnified frames.

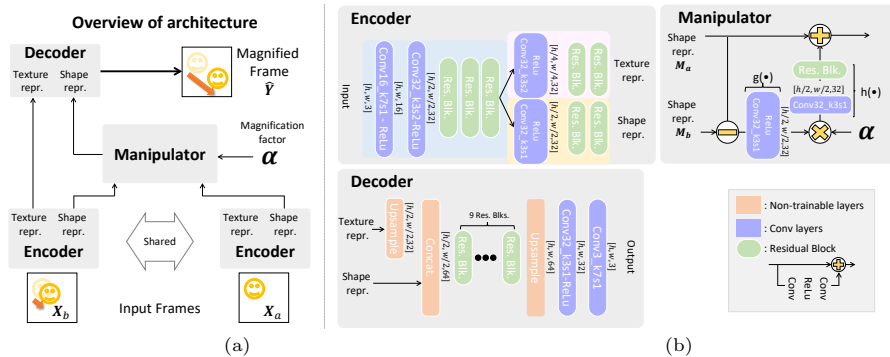

**Fig. 2. Our network architecture.** (a) Overview of the architecture. Our network consists of 3 main parts: the encoder, the manipulator, and the decoder. During training, the inputs to the network are two video frames, $(\mathbf{X}_a, \mathbf{X}_b)$, with a magnification factor $\alpha$, and the output is the magnified frame $\hat{\mathbf{Y}}$. (b) Detailed diagram for each part. `Conv`$\langle c \rangle$`_k`$\langle k \rangle$`_s`$\langle s \rangle$ denotes a convolutional layer of $c$ channels, $k \times k$ kernel size, and stride $s$.

Our encoder and decoder are fully convolutional, which enables them to work on any resolution [16]. They use residual blocks to generate high-quality output [23]. To reduce memory footprint and increase the receptive field size, we downsample the activation by $2\times$ at the beginning of the encoder, and upsample it at the end of the decoder. We downsample with the strided convolution [22], and we use nearest-neighbor upsampling followed by a convolution layer to avoid checkerboard artifacts [20]. We experimentally found that three $3 \times 3$ residual blocks in the encoder and nine in the decoder generally yield good results.

While Eq. (1) suggests no intensity change (constant $f(\cdot)$), this is not true in general. This causes our network to also magnify intensity changes. To cope with this, we introduce another output from the encoder that represents intensity information ("texture representation" [9]) similar to the amplitude of the steerable pyramid decomposition. This representation reduces undesired intensity magnification as well as noise in the final output. We downsample the representation $2\times$ further because it helps reduce noise. We denote the texture and shape representation outputs of the encoder as $\mathbf{V} = G_{e,texture}(\mathbf{X})$ and $\mathbf{M} = G_{e,shape}(\mathbf{X})$, respectively. During training, we add a regularization loss to separate these two representations, which we will discuss in more detail later.

We want to learn a shape representation $\mathbf{M}$ that is linear with respect to $\delta(\mathbf{x}, t)$. So, our manipulator works by taking the difference between shape representations of two given frames, and directly multiplying a magnification factor to it. That is,

$$G_m(\mathbf{M}_a, \mathbf{M}_b, \alpha) = \mathbf{M}_a + \alpha(\mathbf{M}_b - \mathbf{M}_a). \qquad (2)$$

Linear                                    Non-Linear

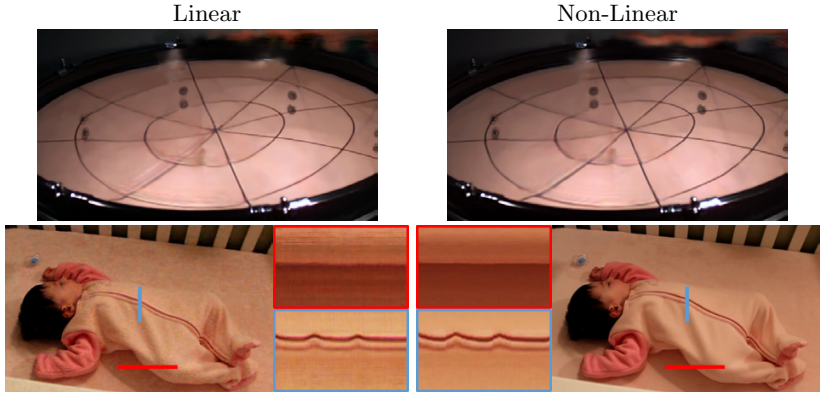

**Fig. 3. Comparison between linear and non-linear manipulators.** While the two manipulators are able to magnify motion, the linear manipulator (left) does blur strong edges (top) sometimes, and is more prone to noise (bottom). Non-linearity in the manipulator reduces this problem (right).

In practice, we found that some non-linearity in the manipulator improves the quality of the result (See Fig. 3). Namely,

$$G_m(\mathbf{M}_a, \mathbf{M}_b, \alpha) = \mathbf{M}_a + h\left(\alpha \cdot g(\mathbf{M}_b - \mathbf{M}_a)\right), \tag{3}$$

where $g(\cdot)$ is represented by a $3 \times 3$ convolution followed by ReLU, and $h(\cdot)$ is a $3 \times 3$ convolution followed by a $3 \times 3$ residual block.

**Loss function.** We train the whole network in an end-to-end manner. We use $l_1$-loss between the network output $\hat{\mathbf{Y}}$ and the ground-truth magnified frame $\mathbf{Y}$. We found no noticeable difference in quality when using more advanced losses, such as the perceptual [8] or the adversarial losses [7]. In order to drive the separation of the texture and the shape representations, we perturbed the intensity of some frames, and expect the texture representations of perturbed frames to be the same, while their shape representation remain unchanged. Specifically, we create perturbed frames $\mathbf{X}'_b$ and $\mathbf{Y}'$, where the prime symbol indicates color perturbation. Then, we impose loses between $\mathbf{V}'_b$ and $\mathbf{V}'_Y$ (perturbed frames), $\mathbf{V}_a$ and $\mathbf{V}_b$ (*un*-perturbed frames), and $\mathbf{M}'_b$ and $\mathbf{M}_b$ (shape of perturbed frames should remain unchanged). We used $l_1$-loss for all regularizations. Therefore, we train the whole network $G$ by minimizing the final loss function $\mathcal{L}_1(\mathbf{Y}, \hat{\mathbf{Y}}) + \lambda(\mathcal{L}_1(\mathbf{V}_a, \mathbf{V}_b) + \mathcal{L}_1(\mathbf{V}'_b, \mathbf{V}'_Y) + \mathcal{L}_1(\mathbf{M}_b, \mathbf{M}'_b))$, where $\lambda$ is the regularization weight (set to 0.1).

**Training.** We use ADAM [11] with $\beta_1 = 0.9$ and $\beta_2 = 0.999$ to minimize the loss with the batch size 4. We set the learning rate to $10^{-4}$ with no weight decay. In order to improve robustness to noise, we add Poisson noise with random strengths whose standard deviation is up to 3 on a $0-255$ scale for a mid-gray pixel.

**Applying 2-frames setting to videos** Since there was no temporal concept during training, our network can be applied as long as the input has two frames.

We consider two different modes where we use different frames as a reference. The `Static` mode uses the 1<sup>st</sup> frame as an anchor, and the `Dynamic` uses the previous frames as a reference, *i.e.* we consider $(\mathbf{X}_{t-1}, \mathbf{X}_t)$ as inputs in the `Dynamic` mode.

Intuitively, the `Static` mode follows the classical definition of motion magnification as defined in Eq. (1), while the `Dynamic` mode magnifies the difference (*velocity*) between consecutive frames. Note that the magnification factor in each case has different meanings, because we are magnifying the motion against a fixed reference, and the velocity respectively. Because there is no temporal filter, undesired motion and noise quickly becomes a problem as the magnification factor increases, and achieving high-quality result is more challenging.

**Temporal operation.** Even though our network has been trained in the 2-frame setting only, we find that the shape representation is linear enough *w.r.t.* the displacement to be compatible with linear temporal filters. Given the shape representation $\mathbf{M}(t)$ of a video (extracted frame-wise), we replace the difference operation with a pixel-wise temporal filter $\mathcal{T}(\cdot)$ across the temporal axis in the manipulator $G_m(\cdot)$. That is, the temporal filtering version of the manipulator, $G_{m,temporal}(\cdot)$, is given by,

$$G_{m,temporal}(\mathbf{M}(t), \alpha) = \mathbf{M}(t) + \alpha \mathcal{T}(\mathbf{M}(t)). \tag{4}$$

The decoder takes the temporally-filtered shape representation and the texture representation of the current frame, and generates temporally filtered motion magnified frames.

### 3.3 Synthetic Training Dataset

Obtaining real motion magnified video pairs is challenging. Therefore, we utilize synthetic data which can be generated in large quantity. However, simulating small motions involves several considerations because any small error will be relatively large. Our dataset is carefully designed and we will later show that the network trained on this data generalizes well to real videos. In this section, we describe considerations we make in generating our dataset.

**Foreground objects and background images.** We utilize real image datasets for their realistic texture. We use $200,000$ images from MS COCO dataset [13] for background, and we use $7,000$ segmented objects the PASCAL VOC dataset [5] for the foreground. As the motion is magnified, filling the occluded area becomes important, so we paste our foreground objects directly onto the background to simulate occlusion effect. Each training sample contains 7 to 15 foreground objects, randomly scaled from its original size. We limit the scaling factor at 2 to avoid blurry texture. The amount and direction of motions of background and each object are also randomized to ensure that the network learns local motions.

**Low contrast texture, global motion, and static scenes.** The training examples described in the previous paragraphs are full of sharp and strong edges where the foreground and background meet. This causes the network to generalize poorly on low contrast textures. To improve generalization in these cases, we add two types of examples: where 1) the background is blurred, and 2) there is only a moving background in the scene to mimic a large object. These improve the performance on large and low contrast objects in real videos.

Small motion can be indistinguishable from noise. We find that including static scenes in the dataset helps the network learn changes that are due to noise only. We add additional two subsets where 1) the scene is completely static, and 2) the background is not moving, but the foreground is moving. With these, our dataset contains a total of 5 parts, each with $20,000$ samples of $384 \times 384$ images. The examples of our dataset can be found in the supplementary material.

**Input motion and amplification factor.** Motion magnification techniques are designed to magnify small motions at high magnifications. The task becomes even harder when the magnified motion is large (*e.g.* > 30 pixels). To ensure the learnability of the task, we carefully parameterize each training example to make sure it is within a defined range. Specifically, we limit the magnification factor $\alpha$ up to 100 and sample the input motion (up to 10 pixels), so that the magnified motion does not exceed 30 pixels.

**Subpixel motion generation.** How subpixel motion manifests depends on demosaicking algorithm and camera sensor pattern. Fortunately, even though our raw images are already demosaicked, they have high enough resolution that they can be downsampled to avoid artifacts from demosaicking. To ensure proper resampling, we reconstruct our image in the continuous domain before applying translation or resizing. We find that our results are not sensitive to the interpolation method used, so we chose bicubic interpolation for the reconstruction. To reduce error that results from translating by a small amount, we first generate our dataset at a higher resolution (where the motion appears larger), then downsample each frame to the desired size. We reduce aliasing when downsampling by applying a Gaussian filter whose kernel is 1 pixel in the destination domain.

Subpixel motion appears as small intensity changes that are often below the 8-bit quantization level. These changes are often rounded away especially for low contrast region. To cope with this, we add uniform quantization noise before quantizing the image. This way, each pixel has a chance of rounding up proportional to its rounding residual (*e.g.*, if a pixel value is 102.3, it will have 30% chance of rounding up).

## 4    Results and Evaluations

In this section, we demonstrate the effectiveness of our proposed network and analyze its intermediate representation to shed light on what it does. We compare qualitatively and quantitatively with the state-of-the-art [26] and show that our network performs better in many aspects. Finally, we discuss limitations of our work. The comparison videos are available in our supplementary material.

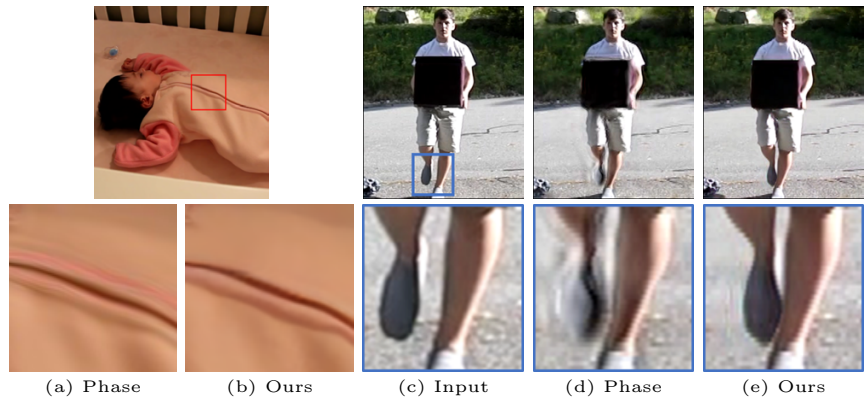

(a) Phase     (b) Ours     (c) Input     (d) Phase     (e) Ours

**Fig. 4. Qualitative comparison.** (a,b) *Baby* sequence (20×). (c,d,e) *Balance* sequence (8×). The phase-based method shows more ringing artifacts and blurring than ours near edges (left) and occlusion boundaries (right).

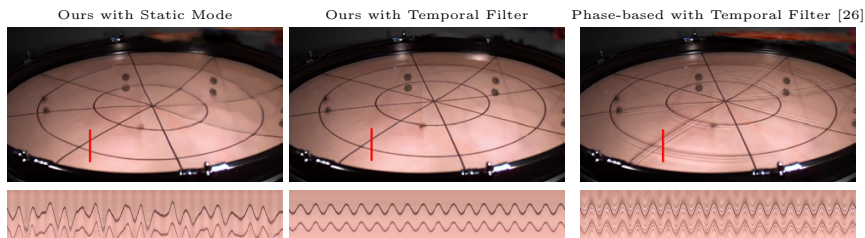

**Fig. 5. Temporal filter reduces artifacts.** Our method benefits from applying temporal filters (middle); blurring artifacts are reduced. Nonetheless, even without temporal filters (left), our method still preserves edges better than the phase-based method (right), which shows severe ringing artifacts.

### 4.1 Comparison with the State-of-the-Art

In this section, we compare our method with the state of the art. Because the Riesz pyramid [27] gives similar results as the steerable pyramids [26], we focus our comparison on the steerable pyramid. We perform both qualitative and quantitative evaluation as follows. All results in this section were processed with temporal filters unless otherwise noted.

**Qualitative comparison** Our method preserves edges well, and has fewer ringing artifacts. Fig. 4 shows a comparison of the *balance* and the *baby* sequences, which are temporally filtered and magnified 10× and 20× respectively. The phase-based method shows significant ringing artifact, while ours is nearly artifact-free. This is because our representation is trained end-to-end from example motion, whereas the phase-based method relies on hand-designed multi-scale representation, which cannot handle strong edges well.

**The effect of temporal filters** Our method was not trained using temporal filters, so using the filters to select motion may lead to incorrect results. To test this, we consider the *guitar* sequence, which shows strings vibrating at different frequencies. Fig. 7 shows the 25× magnification results on the *guitar* sequence using different temporal filters. The strings were correctly selected by each temporal filter, which shows that the temporal filters work correctly with our representation.

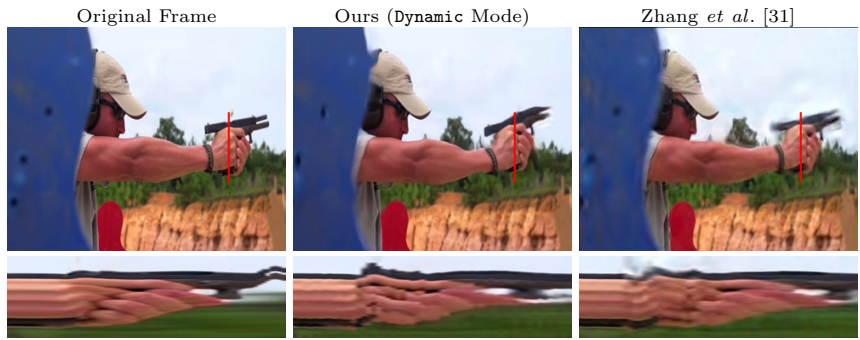

**Fig. 6. Applying our network in 2-frame settings.** We compare our network applied in `dynamic` mode to acceleration magnification [31]. Because [31] is based on the complex steerable pyramid, their result suffers from ringing artifacts and blurring.

Temporal processing can improve the quality of our result, because it prevents our network from magnifying unwanted motion. Fig. 5 shows a comparison on the *drum* sequence. The temporal filter reduces blurring artifacts present when we magnify using two frames (`static` mode). However, even without the use of the temporal filter, our method still preserves edges well, and show no ringing artifacts. In contrast, the phase-based method shows significant ringing artifacts even when the temporal filter is applied.

**Two-frames setting results** Applying our network with two-frames input corresponds best to its training. We consider magnifying consecutive frames using our network (`dynamic` mode), and compare the result with Zhang *et al.* [31]. Fig. 6 shows the result of *gun* sequence, where we apply our network in the `dynamic` mode without a temporal filter. As before, our result is nearly artifact free, while Zhang *et al.* suffers from ringing artifacts and excessive blurring, because their method is also based on the complex steerable pyramid [26]. Note that our magnification factor in the `dynamic` mode may have a different meaning to that of Zhang *et al.*, but we found that for this particular sequence, using the same magnification factor (8×) produces a magnified motion which has roughly the same size.

**Quantitative Analysis.** The strength of motion magnification techniques lies in its ability to visualize sub-pixel motion at high magnification factors, while being resilient to noise. To quantify these strengths and understand the limit of our method, we quantitatively evaluate our method and compare it with the phase-based method on various factors. We want to focus on comparing the representation and not temporal processing, so we generate synthetic examples whose motion is a single-frequency sinusoid and use a temporal filter that has

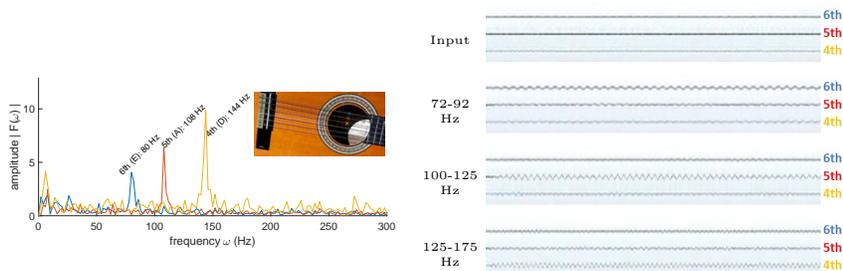

**Fig. 7. Temporal filtering at different frequency bands.** (Left) Intensity signal over the pixel on each string. (Right) $y$-$t$ plot of the result using different temporal filters. Our representation is linear enough to be compatible with temporal filters. The strings from top to bottom correspond to the 6-th to 4-th strings. Each string vibrates at different frequencies, which are correctly selected by corresponding temporal filters. For visualization purpose, we invert the color of the $y - t$ slices.

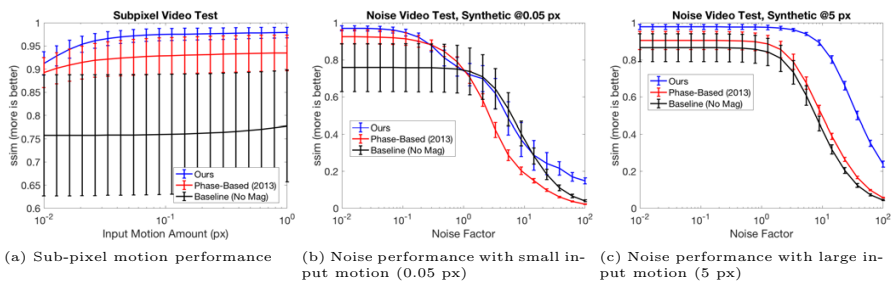

(a) Sub-pixel motion performance

(b) Noise performance with small input motion (0.05 pxl)

(c) Noise performance with large input motion (5 px)

**Fig. 8. Quantitative analysis**. (a) Subpixel test, our network performs well down to 0.01 pixels, and is consistently better than the phase-based [26]. (b,c) Noise tests at different levels of input motion. Our network's performance stays high and is consistently better than the phase-based whose performance drops to the baseline level as the noise factor exceeds 1. Our performance in (b) is worse than (c) because the motion is smaller, which is expected because a smaller motion is harder to be distinguished from noise.

wide passband.[4] Because our network was trained without the temporal filter, we test our method without the temporal filter, but we use temporal filters with the phase-based method. We summarize the results in Fig. 8 and its parameter ranges in the supplementary material.

For the subpixel motion test, we generate synthetic data having foreground input motion ranging from 0.01 to 1 pixel. We vary the magnification factor $\alpha$ such that the magnified motion is 10 pixels. No noise was added. Additionally, we move the background for the same amount of motion but in a different direction to all foreground objects. This ensures that no method could do well by simply replicating the background.

---

[4] Our motion is 3Hz at 30 fps, and the temporal filter used is a 30-tap FIR with a passband between 0.5 - 7.5Hz.

In the noise test, we fixed the amount of input motion and magnification factor and added noise to the input frames. We do not move background in this case. To simulate photon noise, we create a noise image whose variance equals the value of each pixel in the original image. A multiplicative noise factor controls the final strength of noise image to be added.

Because the magnified motion is not very large (10 pixels), the input and the output magnified frames could be largely similar. We also calculate the SSIM between the input and output frames as a baseline reference in addition to the phase-based method.

In all tests, our method performs better than the phase-based method. As Fig. 8-(a) shows, our sub-pixel performance remains high all the way down to 0.01 pixels, and it exceeds 1 standard deviation of the phase-based performance as the motion increase above 0.02 pixels. Interestingly, despite being trained only up to $100\times$ magnification, the network performs considerably well at the smallest input motion (0.01), where magnification factor reaches $1,000\times$. This suggests that our network are more limited by the amount of output motion it needs to generate, rather than the magnification factors it was given.

Fig. 8-(b,c) show the test results under noisy conditions with different amounts of input motion. In all cases, the performance of our method is consistently higher than that of the phase-based method, which quickly drops to the level of the baseline as the noise factor increase above 1.0. Comparing across different input motion, our performance degrades faster as the input motion becomes smaller (See Fig. 8-(b,c)). This is expected because when the motion is small, it becomes harder to distinguish actual motion from noise. Some video outputs from these tests are included in the supplementary material.

## 4.2   Physical Accuracy of Our Method

In nearly all of our *real* test videos, the resulting motions produced by our method have similar magnitude as, and are in phase with, the motions produced by [26] (see Fig. 1, and the supplementary videos). This shows that our method is at least as physically accurate as the phase-based method, while exhibiting fewer artifacts.

We also obtained the hammer sequence from the authors of [26], where accelerometer measurement was available. We integrated twice the accelerometer signal and used a zero-phase high-pass filter to remove drifts. As Fig. 10 shows, the resulting signal (blue line) matches up well with our $10\times$ magnified (without temporal filter) result, suggesting that our method is physically accurate.

## 4.3   Visualizing Network Activation

Deep neural networks achieve high performance in a wide variety of vision tasks, but their inner working is still largely unknown [2]. In this section, we analyze our network to understand what it does, and show that it extracts relevant information to the task. We analyze the response of the encoder, by approximating it

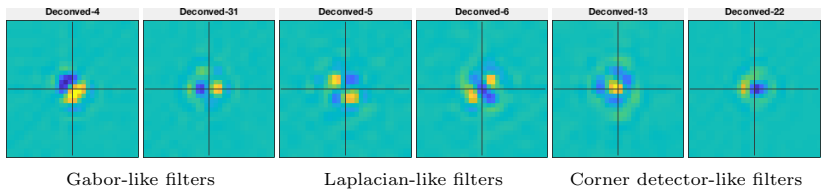

Gabor-like filters          Laplacian-like filters          Corner detector-like filters as a linear system. We pass several test images through the encoder, and calculate the average impulse responses across the images. Fig. 9 shows the samples of the linear kernel approximation of the encoder's shape response. Many of these responses resemble Gabor filters and Laplacian filters, which suggests that our network learns to extract similar information as done by the complex steerable filters [26]. By contrast, the texture kernel responses show many blurring kernels.

**Fig. 9. Approximate shape encoder kernel.** We approximate our (non-linear) spatial encoder as linear convolution kernels and show top-8 by approximation error. These kernels resemble directional edge detector (left), Laplacian operator (middle), and corner detector-like (right).

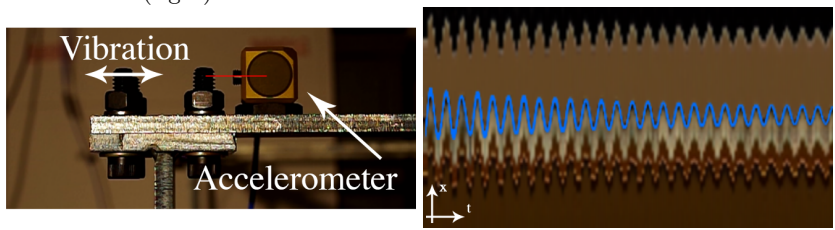

**Fig. 10. Physical accuracy of our method** Comparison between our magnified output and the twice-integrated accelerometer measurement (blue line). Our result and the accelerometer signal match closely.

### 4.4    Limitations

While our network performs well in the 2-frame setting, its performance degrades with temporal filters when the magnification factor is high and motion is small. Fig. 11 shows an example frame of temporally-filtered magnified synthetic videos with increasing the magnification factor. As the magnification factor increases, blurring becomes prominent, and strong color artifacts appear as the magnification factor exceeds what the network was trained on.

In some real videos, our method with temporal filter appears to be blind to very small motions. This results in patchy magnification where some patches get occasionally magnified when their motions are large enough for the network to see. Fig. 12 shows our magnification results of the *eye* sequence compared to that of the phase-based method [26]. Our magnification result shows little motion, except on a few occasions, while the phase-based method reveals a richer motion of the iris. We expect to see some artifact on our network running with temporal filters, because it was not what it was trained on. However, this limits its usefulness in cases where the temporal filter is essential to selecting small

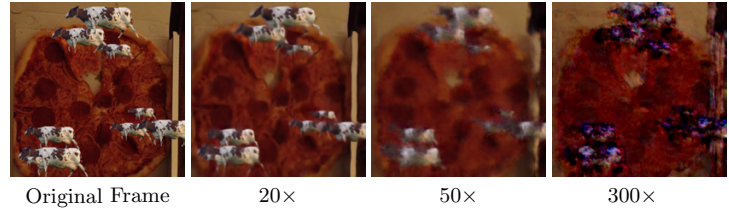

Original Frame      20×      50×      300×

**Fig. 11. Temporal filtered result at high magnification.** Our technique works well with temporal filter only at lower magnification factors. The quality degrades as the magnification factor increases beyond 20×.

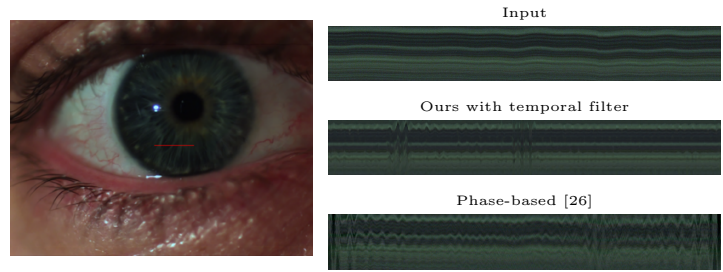

**Fig. 12. One of our failure cases.** Our method is not fully compatible with the temporal filter. This *eye* sequence has a small motion that requires a temporal filter to extract. Our method is blind to this motion and produces a relatively still motion, while the phase-based method is able to reveal it.

motion of interest. Improving compatibility with the temporal filter will be an important direction for future work.

## 5   Conclusion

Current motion magnification techniques are based on hand-designed filters, and are prone to noise and excessive blurring. We present a new learning-based motion magnification method that seeks to learn the filters directly from data. We simplify training by using the two-frames input setting to make it tractable. We generate a set of carefully designed synthetic data that captures aspects of small motion well. Despite these simplifications, we show that our network performs well, and has less edge artifact and better noise characteristics than the state of the arts. Our method is compatible with temporal filters, and yielded good results up to a moderate magnification factor. Improving compatibility with temporal filters so that it works at higher magnification is an important direction for future work.

**Acknowledgment.** The authors would like to thank Qatar Computing Research Institute and Toyota Research Institute for their generous support of this project. Changil Kim was supported by a Swiss National Science Foundation fellowship P2EZP2 168785.

# References

1. Balakrishnan, G., Durand, F., Guttag, J.: Detecting pulse from head motions in video. In: IEEE Conf. on Comput. Vis. and Pattern Recognit. (2013)
2. Bau, D., Zhou, B., Khosla, A., Oliva, A., Torralba, A.: Network dissection: Quantifying interpretability of deep visual representations. In: IEEE Conf. on Comput. Vis. and Pattern Recognit. (2017)
3. Cha, Y.J., Chen, J., Büyüköztürk, O.: Output-only computer vision based damage detection using phase-based optical flow and unscented kalman filters. Engineering Structures **132**, 300–313 (2017)
4. Elgharib, M.A., Hefeeda, M., Durand, F., Freeman, W.T.: Video magnification in presence of large motions. In: IEEE Conf. on Comput. Vis. and Pattern Recognit. (2015)
5. Everingham, M., Van Gool, L., Williams, C.K.I., Winn, J., Zisserman, A.: The pascal visual object classes (voc) challenge. Int. J. of Comput. Vis. **88**(2), 303–338 (Jun 2010)
6. Freeman, W.T., Adelson, E.H.: The design and use of steerable filters. IEEE Trans. Pattern Anal. Mach. Intell. **13**(9), 891–906 (1991)
7. Isola, P., Zhu, J.Y., Zhou, T., Efros, A.A.: Image-to-image translation with conditional adversarial networks. In: IEEE Conf. on Comput. Vis. and Pattern Recognit. (2017)
8. Johnson, J., Alahi, A., Fei-Fei, L.: Perceptual losses for real-time style transfer and super-resolution. In: Eur. Conf. on Comput. Vis. Springer (2016)
9. Jones, M.J., Poggio, T.: Multidimensional morphable models: A framework for representing and matching object classes. Int. J. of Comput. Vis. **29**(2), 107–131 (1998)
10. Kalantari, N.K., Wang, T.C., Ramamoorthi, R.: Learning-based view synthesis for light field cameras. ACM Trans. Graph. (SIGGRAPH Asia) **35**(6), 193–10 (2016)
11. Kingma, D.P., Ba, J.: Adam: A method for stochastic optimization. arXiv preprint arXiv:1412.6980 (2014)
12. Liao, Z., Joshi, N., Hoppe, H.: Automated video looping with progressive dynamism. ACM Trans. Graph. (SIGGRAPH) **32**(4), 77 (2013)
13. Lin, T.Y., Maire, M., Belongie, S., Hays, J., Perona, P., Ramanan, D., Dollár, P., Zitnick, C.L.: Microsoft coco: Common objects in context. In: Eur. Conf. on Comput. Vis. Springer (2014)
14. Liu, C., Torralba, A., Freeman, W.T., Durand, F., Adelson, E.H.: Motion magnification. ACM Trans. Graph. (SIGGRAPH) **24**(3), 519–526 (2005)
15. Liu, Z., Yeh, R.A., Tang, X., Liu, Y., Agarwala, A.: Video Frame Synthesis using Deep Voxel Flow. In: IEEE Int. Conf. on Comput. Vis. (2017)
16. Long, J., Shelhamer, E., Darrell, T.: Fully convolutional networks for semantic segmentation. In: IEEE Conf. on Comput. Vis. and Pattern Recognit. (2015)
17. Mathieu, M., Couprie, C., LeCun, Y.: Deep multi-scale video prediction beyond mean square error. Int. Conf. on Learn. Representations (2016)
18. Niklaus, S., Mai, L., Liu, F.: Video Frame Interpolation via Adaptive Convolution. IEEE Conf. on Comput. Vis. and Pattern Recognit. (2017)
19. Niklaus, S., Mai, L., Liu, F.: Video Frame Interpolation via Adaptive Separable Convolution. In: IEEE Int. Conf. on Comput. Vis. (2017)
20. Odena, A., Dumoulin, V., Olah, C.: Deconvolution and checkerboard artifacts. Distill **1**(10), e3 (2016)

21. Oh, T.H., Joo, K., Joshi, N., Wang, B., Kweon, I.S., Kang, S.B.: Personalized cinemagraphs using semantic understanding and collaborative learning. In: IEEE Int. Conf. on Comput. Vis. (2017)

22. Radford, A., Metz, L., Chintala, S.: Unsupervised representation learning with deep convolutional generative adversarial networks. arXiv preprint arXiv:1511.06434 (2015)

23. Sajjadi, M.S., Schölkopf, B., Hirsch, M.: EnhanceNet: Single image super-resolution through automated texture synthesis. In: IEEE Int. Conf. on Comput. Vis. (2017)

24. Srivastava, N., Mansimov, E., Salakhudinov, R.: Unsupervised learning of video representations using lstms. In: Int. Conf. on Mach. Learn. (2015)

25. Villegas, R., Yang, J., Hong, S., Lin, X., Lee, H.: Decomposing motion and content for natural video sequence prediction. In: Int. Conf. on Learn. Representations (2017)

26. Wadhwa, N., Rubinstein, M., Durand, F., Freeman, W.T.: Phase-based video motion processing. ACM Trans. Graph. (SIGGRAPH) **32**(4),  80 (2013)

27. Wadhwa, N., Rubinstein, M., Durand, F., Freeman, W.T.: Riesz pyramids for fast phase-based video magnification. In: IEEE Int. Conf. on Comput. Photogr. (2014)

28. Wang, T., Zhu, J., Kalantari, N.K., Efros, A.A., Ramamoorthi, R.: Light field video capture using a learning-based hybrid imaging system. ACM Trans. Graph. (SIGGRAPH) **36**(4), 133:1–133:13 (2017)

29. Wilburn, B., Joshi, N., Vaish, V., Talvala, E.V., Antunez, E., Barth, A., Adams, A., Horowitz, M., Levoy, M.: High performance imaging using large camera arrays. ACM Transactions on Graphics (TOG) **24**(3), 765–776 (2005)

30. Wu, H.Y., Rubinstein, M., Shih, E., Guttag, J., Durand, F., Freeman, W.: Eulerian video magnification for revealing subtle changes in the world. ACM Trans. Graph. (SIGGRAPH) **31**(4), 65–8 (2012)

31. Zhang, Y., Pintea, S.L., van Gemert, J.C.: Video Acceleration Magnification. In: IEEE Conf. on Comput. Vis. and Pattern Recognit. (2017)

# Supplementary Material:
# Learning-based Video Motion Magnification

Tae-Hyun Oh[1*], Ronnachai Jaroensri[1*], Changil Kim[1], Mohamed Elgharib[2],
Frédo Durand[1], William T. Freeman[1,3], and Wojciech Matusik[1]

[1] MIT CSAIL, Cambridge, MA, USA
[2] HBKU QCRI, Doha, Qatar
[3] Google Research
{taehyun, tiam}@csail.mit.edu

## Summary of Contents

This is a part of the supplementary material. The contents of this supplementary material include the additional experiment results with descriptions and parameter setups, which have not been shown in the main paper due to the space limit.

The supplementary video[4] contains comparisons with other methods [30, 26, 27, 31] and baselines, self-evaluations and other applications.

## A.1    Parameters for Example Videos

In Table A.1, we specify parameters used in the experiments of the main paper such as magnification factors, and temporal filters. The parameters of examples in the supplementary video are self-contained if specified. For FIR filter, we chose the number of taps equal to the number of video frames in that sequence, and apply the filter in the frequency domain. For all other filters, we apply them in the time domain.

## A.2    Additional Experiments

This section describes the detail description of the dataset (in Sec. A.2.1), and presents additional results. We present an example for the characteristics of `Static` mode that magnifies broad frequency bands in Sec. A.2.2, qualitative comparison in Sec. A.2.3, quantitative analysis in Sec. A.2.4, visualization analysis examples in Sec. A.2.5, descriptions for other applications in Sec. A.2.6 and the supplementary video content in Sec. A.2.7.

---

[*] These authors contributed equally.
[4] The supplementary video can be found in https://youtu.be/GrMLeEcSNzY.

| Sequence Name | Magnification Factor | Temporal Band | Sampling Rate (fps) | Temporal Filter |
|---|---|---|---|---|
| crane | $75\times$ | $0.2 - 0.25$ Hz | 24 | FIR |
| balance | $10\times$ | $1 - 8$ Hz | 300 | $2^{nd}$ order Butterworth |
| throat | $100\times$ | $90 - 110$ Hz | 1900 | FIR |
| baby | $20\times$ | See Temporal Filter Column | 30 | Difference of IIR, same as [27] |
| tree (high freq band) | $25\times$ | $1.5 - 2$ Hz | 60 | FIR |
| tree (low freq band) | $25\times$ | $0.5 - 1$ Hz | 60 | FIR |
| camera | $75\times$ | $36 - 62$ Hz | 300 | $2^{nd}$ order Butterworth |
| eye | $75\times$ | $30 - 50$ Hz | 500 | FIR |
| gun | $8\times$ | N/A | N/A | `Dynamic` Mode |
| cat-toy | $7\times$ | N/A | N/A | `Dynamic` Mode |
| drone | $10\times$ | N/A | N/A | `Dynamic` Mode |
| hot-coffee | $3\times$ | N/A | N/A | `Dynamic` Mode |
| drum | $10\times$ | $74 - 78$ Hz | 1900 | FIR |
| drum | $10\times$ | N/A | N/A | `Static` Mode |
| guitar | $25\times$ | $72 - 92$ Hz | 600 | $2^{nd}$ order Butterworth |
| guitar | $10\times$ | N/A | N/A | `Static` Mode |

**Table A.1.** Parameters of our results

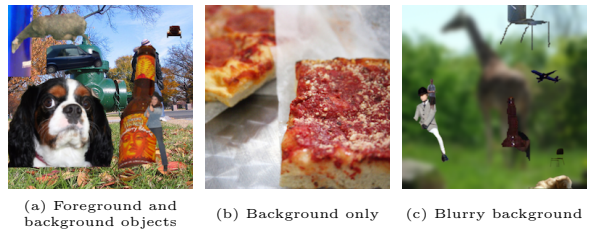

(a) Foreground and background objects    (b) Background only    (c) Blurry background

**Fig. A.1. Sample data.** Example frames from our dataset. (a) Our data consists of pasted foreground objects using segmentation from PASCAL VOC [5] and background from MS COCO dataset [13]. (b) To ensure our network learns global motion, the second part of our dataset only has background moving. (c) To ensure low contrast texture is well-represented, we include data with the background blurred. We add another two parts with the same specification as (b) and (c) with background motion removed so that the network learns the changes that are due to noise only.

### A.2.1   Detail dataset Information.

In the synthetic training data generation, we use two datasets: the MS COCO [13], and the PASCAL VOC segmentation dataset [5]. Fig. A.1 shows example frames from our dataset. The license conditions are as follows: the annotations and images of the MS COCO are under Creative Commons Attribution 4.0 license and

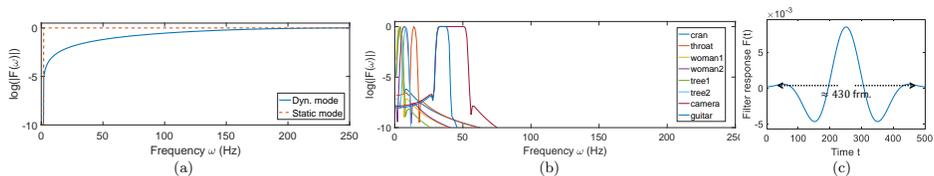

**Fig. A.2. Comparison on temporal operators.** The plots visualize log frequency responses of the subtraction operation for `Static` and `Dynamic` modes (a) and temporal filter examples (b) used in [30, 26, 27]. (c) temporal domain visualization of a low-freq. band-pass filter example (sampling rate: 60Hz, cut-off: [0.35, 0.71]Hz) used in [26], where the filter lies across near 450 frames. For visual comparison purpose, the frequency domain range is resampled to have the same range comparable between (a) and (b).

Flickr terms of use, respectively, and the PASCAL VOC is under Flickr terms of use and MSR Cambridge License (RTF).

All the real video examples used for comparisons come from either [30, 27, 26, 4, 31]; otherwise our own captured data is used.

### A.2.2    Comparison of temporal operations

Our network is trained on the two frames setting, but we also have shown that multi-frame linear temporal filtering is compatible to some extent. We discuss the difference of temporal operation characteristics for a reference purpose. Fig. A.2 shows the frequency response characteristics of different temporal operations.

Also, as shown in Fig. A.2, the area-under-covers of (a) are far broader than (b), *i.e.*, these modes magnify broader frequency ranges comparing to band-pass filtering in (b). This implies as follows. First, as the same amount of magnification factors are increased, overall energy increment of ours is even larger than the previous work; hence, the magnification factors are not directly comparable across modes, and this fact needs to be taken into account when the results are compared. Second, since noise (high) frequency band is also involved in, the training and working on the regime (a) would require more robust representation and synthesis power than the regime (b); thus, the regimes of utilizing only two frames are much challenging. For temporal filtering, magnifying a low-frequency band needs to take into account a long history of frames as shown in Fig. A.2-(c), while given an anchor frame, `Static` mode can magnify broad ranges of motion only with two frames, which is memory efficient.

We note that the subtraction operation at test phase is not claimed as the best, but they have trade-offs. The multi-frame based temporal filtering [30, 26] has its own merits: selectivity of a frequency band of interest, and noise frequency band suppression as shown in Fig. A.2-(b). Since our network is trained only on (a), it is hard to expect any generalization for other temporal operations. Surprisingly, at the test stage, we observed that our representation works favorably well with replacing the subtraction operation in the manipulator by the temporal filtering. Thus, we also conduct experiments with the subtraction operation to assess and compare the performance of our representation with the competing ones, *i.e.*, phase representations from complex steerable filters [26] and Riesz transformation [27] (see the supplementary video). Also, we compare our method with the competing methods equipped with temporal filtering.

Fig. A.3 shows its example that `Static` mode with our method magnifies the motion of all three strings whose frequencies are different. This shows that our method magnifies true motions, but not hallucinating.

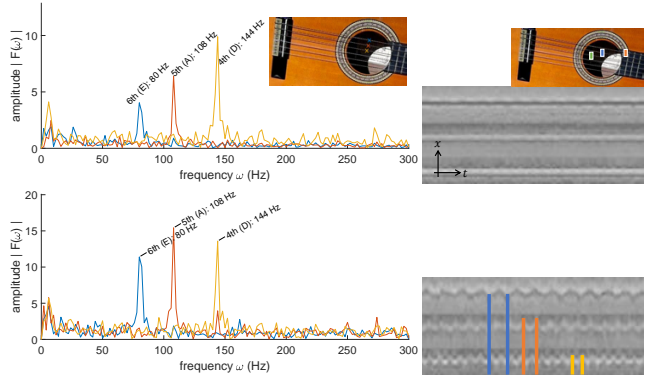

**Fig. A.3. Broad frequency magnification (Guitar sequence).** The `Static` mode with our method does not alter input frequency, as well as magnifies overall the frequency. The estimated frequencies of $4, 5, 6$-th strings are $\{137.14, 109.10, 80\}$-Hz, which is very close values to ideal frequencies, $\{144, 108, 80\}$-Hz. Color lines on the slice view are the samples of the period measure to estimate frequency.

### A.2.3    Additional qualitative Comparisons.

Various qualitative results can be found in the supplementary video. *Due to the characteristic of the problem, we strongly recommend to refer to the supplementary video for complete comparison and evaluation.*

We also present another comparison with other methods in Fig. A.4, where we replace the all the temporal operation to be `Dynamic` mode, so that we can compare the capability of the representation and synthesis of each method under the same setting.

### A.2.4    Additional Quantitative Comparisons.

We summarize the test parameters used in all quantitative results in Table A.2. In Fig. A.5, we show additional quantitative evaluations for completeness: noise performance on real data and input motion range test. To test the noise performance on real examples in (a), we grab three consecutive frames from 17 high-speed videos, and approximate their motions as linear. That is, the third

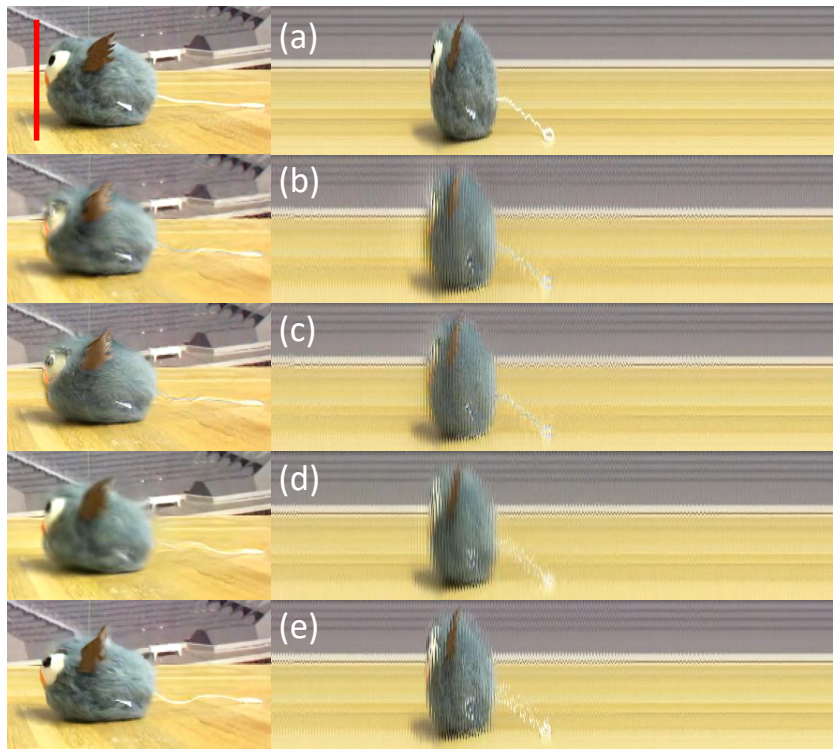

**Fig. A.4. Slice view comparison with other methods, `Dynamic mode`.** We compare original inputs (a) for reference, Wadhwa *et al.* [26] (b), Wadhwa *et al.* [27] (c), Zhang *et al.* [31] (d) and the proposed method (e). Even though there are large translational motions, only our method shows clear boundaries with proper magnified motion in slice view, while the other methods have blurry textures due to artifacts.

**Table A.2. Summary of parameter for each quantitative test.**

| Test<br>Param. | Magnification Factor | Motion (px.) | Magnified Motion (px.) | Noise | Background Motion |
|---|---|---|---|---|---|
| Sub-pixel | Capped at 100 | 0.01 - 1 | Capped at 10 | No | Yes |
| Noise | 5.0 | 2.0 | 10.0 | 0.01 - 100 | No |
| Noise on Real Image | 2.0 (estimate) | - | - | 0.01 - 100 | No |
| Magnification Factor | 1 - 1,000 | 0.01 - 2.0 | Capped at 10 | No | Yes |
| Input Motion | Capped at 5.0 | 1 - 10 | Capped at 10 | No | Yes |
| Magnified Motion | Capped at 5.0 | 1 - 10 | 1 - 30 | No | Yes |

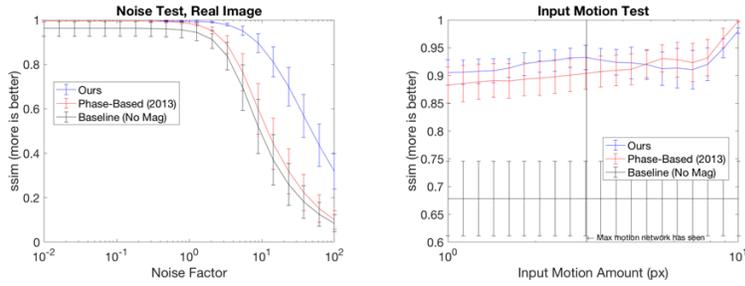

(a) Noise performance in real images

(b) Input motion test. Because magnified motion is capped at 10px, magnification factor approaches 1 towards the end, which explain the performance.

**Fig. A.5. Additional Quantitative Analysis**.

frame will have twice the amount of motion as the second frame. We synthetically add noise as described before.

To see the extreme behavior against noise, we qualitatively compare the cases in Fig. A.6. Our method tends to preserve detail better (notice the car) rather than blurring, and have a better structure preserving quality in the output than the phase-based one. Since our filters are directly learned in a data-driven way, we can easily augment the data to induce desired properties.

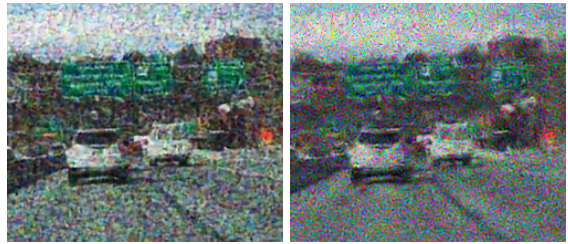

(a)                    (b)

**Fig. A.6. Example output from extreme noise test on real images (Noise factor = 9)**. (a) Our method preserves the detail better and results in less noise than (b) the phase-based method [26].

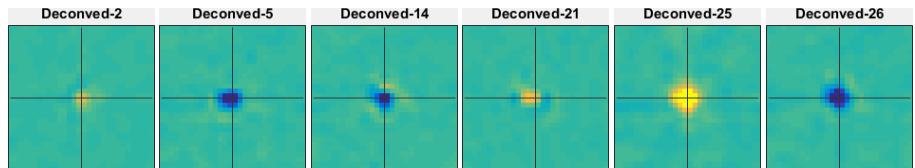

**Fig. A.7. Approximate texture encoder kernel.** We approximate our (non-linear) spatial encoder as linear convolution kernels. Most of the kernel approximations resemble narrow blur kernels. This differs from the shape representation kernel (Fig. 9).

### A.2.5    Additional Network Visualization for Analysis

We present the additional visualization. Further observation can be obtained by directly visualizing each unit (channel) with a simple example. Interestingly, we observed that many activation patterns could be categorized into four cases shown in Fig. A.8. Motivated by the recent network interpretation method [2], we use an example based interpretation. We use a simple synthetic data with apparently simple colors with several motions (see Fig. A.8-(a)). We compare the texture and shape representation at early and later iteration of training. Fig. A.8 shows that the texture representations tend to be direction invariant and color sensitive, while the shape representation tend to be color invariant and direction sensitive.

Specifically, we inspect the transitions by counting the number of directional invariant detectors in texture representation, and numbers of color invariant detectors in shape representation at 5k and 30k iterations by subjects with referring to the references in the left. The ratios of directional invariant and color invariant detectors are $\frac{22}{32}$ (68.8%)$\rightarrow\frac{15}{32}$ (46.9%), and $\frac{26}{64}$ (40.6%)$\rightarrow\frac{37}{64}$ (57.8%), respectively, according to training. Since this behavior is learned from data, it may suggest that the steerable filter (directional) for shape representation in [26] was a good choice for the magnification task.

In Fig. A.8-(j,k), we sample the representation randomly from all units, where it shows that detectors dominant to either color or directional information are frequently observed in respective representations.

We additionally visualize the kernel approximation of the texture representation as done in Sec. 4.3 of the main paper. The kernels of the texture representation (Fig. A.7) are different from those of the shape representation. They seem to resemble blurring kernels (or delta function).

In Fig. A.9-(a,b), we compare the activations of the compensation part, *i.e.*, $h(\alpha(g(\mathbf{A}_2 - \mathbf{A}_1)))$, with two different magnification factors $\alpha$ for the same input motion. It shows that, according to $\alpha$, the compensation not only scales activations but also spatially propagates around. This suggests that our representation acts as a proxy of motion information that is able to be interpreted and manipulated. Lastly, observing activations of the compensation part in Fig. A.9 and the fusion layer in Fig. A.10, the network seems to learn how to compensate discrepancy introduced by motion manipulation, rather than explicitly moving pixels. These suggest that our method learns a representation closely related to Eulerian motion representation rather than Lagrangian one.

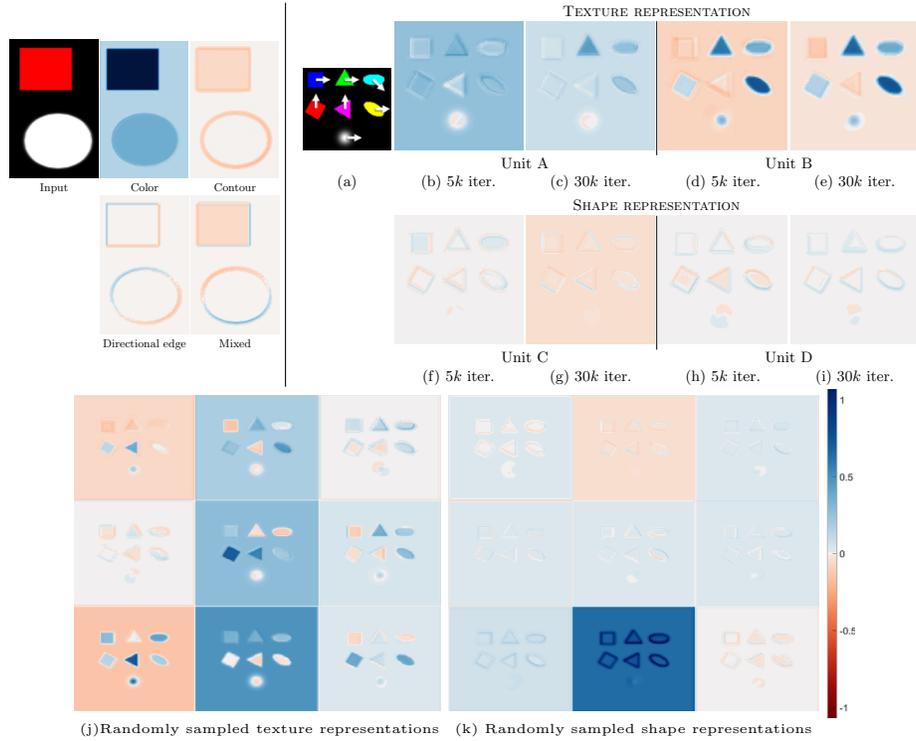

**Fig. A.8. Visualization of representations from the encoder.** We observed that many texture and shape activations could be mainly categorized into four types of dominant characteristics: color, contour, directional edge and mixed detectors. With this observation, we only exploit simple color and directional edge property for maintaining interpretability. (a) Synthetic data used for analysis, where input motion displacements are {0.2, 0.5, 0.8, 1.1, 1.4, 1.7, 2.0} pixels in a raster scan order. (b–i) We probe the hidden units that behave like a sort of invariant detector at 5k and 30k iterations. The texture representation tends to capture color properties, while the shape one tends to capture edge and boundary. Especially, Unit A evolves to be a green color dominant detector (refer to the color of (a)) with losing directional responses. Unit D evolves from a red color plus directional edge detector to color invariant edge detector. (j,k) Randomly sampled representations at TEXTURE REPR. and SHAPE REPR. in Fig. 2, respectively.

### A.2.6    Applications

To show the versatility of our learned representation, we additionally present potential applications, view/frame interpolation. and dynamic cinemagraph. Note that we directly use the network trained on our synthetic data for motion magnification, and we have never re-trained or fine-tuned the network for the specific applications. The results can be found in the supplementary video.

For the view/frame interpolation applications, we use our network to generate intermediate frames between two input frames by changing magnification factor in a range of [0,1]. For the view interpolation example, we generate and insert 25 intermediate frames given two input images. For the frame interpolation example, we generate five frames between every two consecutive frames, *i.e.*, 5× temporal interpolation. We use the light field camera dataset [29] in these applications.

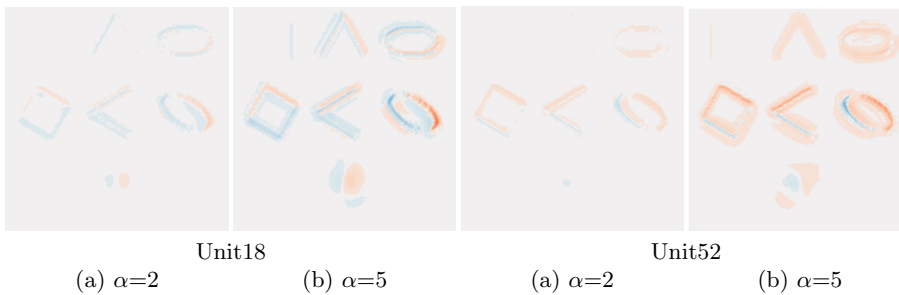

|  | Unit18 |  | Unit52 |
|---|---|---|---|
| (a) $\alpha=2$ | (b) $\alpha=5$ | (a) $\alpha=2$ | (b) $\alpha=5$ |

**Fig. A.9. Visualization of activations of $h(\alpha(g(\mathbf{A}_2 - \mathbf{A}_1)))$.** Comparing (a) to (b), as the magnification factors are increased, the manipulator layer not only increases the scale of activation, but also propagates the activation around. In this regard, our network seems to learn how to compensate object movements according to resulting motion.

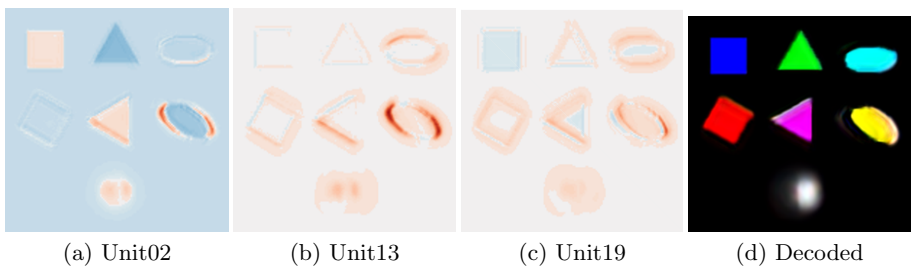

(a) Unit02        (b) Unit13        (c) Unit19        (d) Decoded

**Fig. A.10. Visualization of activations at the first layer residual block in the decoder, Fusion layer.** We show randomly sampled activations fused from texture and shape compensated representations. The representations and decoded frame show that our network compensates discrepancy induced by motion magnification visually, rather than explicit pixel movement. It shows a color synthesis behavior of our network in (d). Around the yellow ellipse in (d), the color values seem to be synthesized to compensate movements in a similar way to inpainting.

For the cinemagraph application [12], we add a dynamism control function by a post-synthesis approach using our method. We can control the magnitude of dynamism in the cinemagraph by leveraging the controllability of magnification factors without retraining. We use the cinemagraph data generated by [21].

### A.2.7    Supplementary Video Content

The content summary of the supplementary video[5] is as follows:

---

[5] The supplementary video can be found in https://youtu.be/GrMLeEcSNzY.

- Qualitative comparison with temporal filter
- Qualitative comparison in 2 frame input setting (static mode, dynamic mode, frequency characteristics comparison)
- Additional analysis
  - Magnification ability according to (sub-pixel) input motion
  - Applying different magnification factors without re-training (including motion attenuation examples)
- Applications: View/Frame interpolation and progressive dynamism effects of cinemagraph.